\pdfoutput=1

\documentclass[11pt]{article}

\usepackage{acl}
\usepackage{pgfplots}
\usepackage{amsmath}
\usepackage{times}
\usepackage{latexsym}
\usepackage{booktabs}
\usepackage[T1]{fontenc}
\usepackage{graphicx}
\graphicspath{ {images/} }

\usepackage[utf8]{inputenc}
\usepackage{float}
\usepackage{graphicx}
\usepackage{microtype}

\usepackage{inconsolata}

\usepackage{tcolorbox}
\tcbuselibrary{skins}

\usepackage{xcolor}

\title{Chain-of-Thought Augmentation with Logit Contrast for Enhanced Reasoning in Language Models}

\author{Jay Shim\thanks{Lead Authors with Equal Contribution. Order chosen randomly.} \hspace{1cm} Grant Kruttschnitt\footnotemark[1] \hspace{1cm} Alyssa Ma \hspace{1cm} Daniel Kim \\
{\bf Benjamin Chek} \hspace{1cm} {\bf Athul Anand} \hspace{1cm} \hspace{1cm}
        {\bf Kevin Zhu}\thanks{Senior Author.} \hspace{1cm} {\bf Sean O'Brien}\footnotemark[2]  \\
        Algoverse AI Research \\
        \texttt{sean@algoverse.us, kevin@algoverse.us}}

\begin{document}
\maketitle
\begin{abstract}
Rapidly increasing model scales coupled with steering methods such as chain-of-thought prompting \citep{Wei2022ChainOT} have led to drastic improvements in language model reasoning.
At the same time, models struggle with compositional generalization and are far from human performance on many reasoning-based benchmarks.
Leveraging the success of chain-of-thought prompting, and also taking inspiration from context-aware decoding (CAD) \citep{shi2023trusting},
we explore input-based contrasting methods to further encourage the type of reasoning induced by chain-of-thought prompting.
While work remains to stabilize these results across datasets and models, the improvements we find warrant further investigation into input-based steering methods for context-aware reasoning.
\end{abstract}

\section{Introduction}
Large Language Models (LLMs) have revolutionized natural language processing by leveraging vast amounts of data to generate human-like text. These models have gained attention for their ability to perform a wide range of language tasks, from text generation and translation to question answering and summarization. One such technique that has propelled the performance of LLMs is Chain-of-Thought (CoT) prompting \citep{Wei2022ChainOT} . By applying CoT prompting, researchers have found substantial improvement in various tasks such as arithmetic and common sense reasoning, surpassing the efficacy of traditional methods focused on scaling model size \citep{wang2023selfconsistency,wei2023chainofthought}. In a similar vein, Context-Aware Decoding (CAD) emerges as another approach to augment LLM performance \citep{liu2021dexperts, li2023contrastive, obrien2023contrastive}, 
prioritizing relevant contextual information over pre-existing knowledge \citep{shi2023trusting}. While these advancements underscore significant progress, an ongoing question persists regarding the efficacy of combining CAD with CoT or with amateur prompts, which typically contain less context.

To address this question, the study aims to investigate how CoT prompts' ability to capture sequential dependencies can be complemented by the discriminative ability of CAD, particularly when the context contradicts prior knowledge \citep{longpre-etal-2021-entity, zhou2023contextfaithful}. In such cases, we aim to ensure that incorporating CoT prompting does not detract from the outputs derived from the provided context. Ultimately, the goal is to create faithful models capable of accurately executing mathematical and reasoning tasks while avoiding the misinterpretation of crucial contextual cues \citep{maynez-etal-2020-faithfulness,pagnoni-etal-2021-understanding}.

\section{Method}
\subsection{Background}
We take a query $\textbf{x}$ which we feed to our language model $\theta$, to generate a response $\textbf{y}$ using greedy decoding. The response is sampled autoregressively from two probability distributions given by query $x$, CoT prompt $\textbf{c}$, and an amateur prompt (\ref{sec:am1}) with less context $\textbf{amt}$ \citep{shi2023trusting}. Using a contrastive strength of $\alpha$, we create a new distribution:

\[
y_t \propto \frac{p_{\theta}(y_t | \mathbf{c, x, y_{<t}})^{\alpha + 1}
}{p_{\theta}(y_t | \mathbf{amt, x, y_{<t}})^{\alpha}
}
\]

We hypothesize that the model can effectively rely on the prompt $\textbf{c}$ when reasoning with the input while simultaneously penalizing incorrect reasoning behaviors and conciseness ($\textbf{amt}$) to provide an accurate output.

\subsection{Context-Aware Decoding}
For each method we utilize two different prompts: an 8-shot CoT prompt which we call an expert prompt, and an amateur prompt. To identify the next logit and thus the next token to add autoregressively, we use the contrastive function below proposed by \citet{shi2023trusting}:
\[
y_t \sim \textrm{softmax} \left[ (1+\alpha) \exp(\text{logit}_{\theta}(y_t | \mathbf{c, x, y_{<t}})) \right.
\]
\[
\left. - \alpha \exp(\text{logit}_{\theta}(y_t | \mathbf{amt, x, y_{<t}})) \right]
\]

The values of $\alpha$ are as follows, $0 \leq \alpha \leq 1$, with higher alpha values indicating stronger reliance on prompt $\textbf{c}$, the CoT prompt. From the adjusted logits, we apply greedy decoding to generate the subsequent token. In our analysis, a single model was used to generate probabilities for both CoT and amateur prompts, but different models may also be used.

\section{Experimental Setup}

\subsection{Datasets}

We evaluated our method on three question-answering task datasets that measure performance based on reasoning abilities: GSM8K \citep{cobbe2021training}, AQuA \citep{ling2017program}, and CommonSenseQA \citep{talmor2019commonsenseqa}.
We provide multiple-choice options in the prompts of CommonSenseQA and AQuA, but do not for GSM8K. GSM8K and AQuA are math datasets, while CommonSenseQA contains common-sense reasoning problems. 

\subsection{Models and Prompting}
We performed our experiments using Phi-1.5 \citep{li2023textbooksneediiphi15}, Mistral 7B \citep{jiang2023mistral7b}, and GPT-3.5 \citep{openai2024gpt4technicalreport} with version 3.5-turbo-0125. The reason we chose these models is that they were varied. The reason we used Phi-1.5 was because it was very low parameters compared to the other two models. We used Mistral, because it had more parameters, so it was more accurate. The reason we used GPT 3.5 was to compare it with the other two open-source models, and because it was more parameters than Mistral. However, our use of GPT 3.5 was limited, because it cost a lot, so we were only able to run baseline and one Amateur on one Alpha.
In each of our experiments, we contrasted the logits obtained from a 8-shot CoT prompt with the amateur. The exemplars in the few-shot prompts are taken from the corresponding dataset used for evaluation. 

To clarify, for each of these, the amateur prompt is different, but has a limitation which should make it worse. We contrast it against the expert prompt, which had the same format across all experiments, as seen below.

\begin{tcolorbox}[title=Expert Prompt (8-shot CoT), colback=white, colframe=black, sharp corners]
Q: \{\textit{question 1}\}\\
A: \{\textit{CoT 1}\} \{\textit{answer 1}\}\\
$\cdots$\\
Q: \{\textit{question 8}\}\\
A: \{\textit{CoT 8}\} \{\textit{answer 8}\}\\
Q: \{\textit{new question}\}\\
A: \\
\end{tcolorbox}


\subsubsection{Expert vs Amateur 1}\label{sec:am1}
In our main method, the amateur is given no context, not even the question. By doing so, we increase the potential of our context-aware decoding and follow the formulation of \citet{sanchez2023stay}'s method. \\ The effectiveness on this method is that this will cause it to hopefully rely on the chain of thought process, and the actual question much more, since the tokens that do rely on it will be much more likelier to occur. 



\begin{tcolorbox}[enhanced, title=Amateur 1 Prompt, colback=white, colframe=black, sharp corners]
A: \\
\end{tcolorbox}

\subsubsection{Expert vs Amateur 2} \label{sec:am2}
In our second amateur, we feed an 8-shot CoT prompt with the questions omitted. This should encourage faithfulness to the question context, as demonstrated in factuality-based tasks by \citep{shi2023trusting}. That is because it would reduce the probability of logits that would happen even if there's no question, and increase the probabilities of those that only appear because of it, making the logits that eventually end in it more likely to rely on the question.\\

\begin{tcolorbox}[enhanced, title=Amateur 2 Prompt, colback=white, colframe=black, sharp corners]
A: \{\textit{CoT answer 1}\}\\
$\cdots$ \\
A: \{\textit{CoT answer 8}\}\\
Q: \{\textit{new question}\}\\
A:
\end{tcolorbox}

\subsubsection{Expert vs Amateur 3} \label{sec:am3}
In the third amateur, we feed an 8-shot prompt with questions and answers, but without CoT reasoning. This should hopefully induce further step-by-step reasoning, under the intuition that a reasoning chain uniquely preferred by a CoT prompted model will exhibit more thorough reasoning. \\

\begin{tcolorbox}[enhanced, title=Amateur 3 Prompt, colback=white, colframe=black, sharp corners]
Q: \{\textit{question 1}\}\\
A: \{\textit{non-CoT answer 1}\}\\
$\cdots$ \\
Q: \{\textit{question 8}\}\\
A: \{\textit{non-CoT answer 8}\}\\
Q: \{\textit{new question}\}\\
A:
\end{tcolorbox}

\subsection{Hyperparameter Search}

Measuring performance on Mistral 7B \citep{jiang2023mistral} using the CommonSenseQA dataset \citep{talmor2019commonsenseqa}, we found an optimal $\alpha$ value of $0.8$ (see ~\autoref{tab:alpha-sweep}).
This is a constant we used across all datasets and models. In addition, we decided to use 0.5, because although it did not have the second highest accuracy, we wanted another alpha away from 0.8. That would allow us to see a more pronounced difference between the alphas.

\begin{table}[H]
    \centering
    \begin{tabular}{|c|c|}
        \toprule
        $\alpha$ & Accuracy \\
        \midrule
        0.5 & 0.500 \\
        0.7 & 0.620 \\
        0.8 & 0.646 \\
        0.9 & 0.59   \\
        \bottomrule 
    \end{tabular}
        
    \caption{Accuracy of different CAD levels while contrasting CoT and Amateur 3 with Mistral on 200 questions of CommonSenseQA}
    \label{tab:alpha-sweep}
\end{table}

\subsection{Baselines}
We first run our datasets through 8-shot CoT prompting, without contrasting with any amateur prompt, setting the benchmark accuracy for the rest of the contrasting experiments. We used the same 8-shot CoT prompt as we did for our experts.




\section{Results}

As seen in \hyperref[sec:Table 2]{Table 2}, CAD improved AQuA's accuracy on Phi by a substantial result, and the same for Mistral on CommonSenseQA. The other accuracy improvements were much more modest, with three percent and below in improvement. The AQuA improvement might not show that much importance, however, because even the improved accuracy still mostly stays at the random rate of guessing.


The CoT Contrast works best with Phi-1.5 on AQuA, while it works best with Mistral 7B on CommonsenseQA. Therefore, we do not find that contrastive CoT increases problem solving performance in all language models for all data.

The reason that we did not put GPT-3.5 into table 2 was because of prohibitive costs of GPT tokens, we could not run the same experiments on GPT that we did on Mistral and Phi.

In Table 3, GSM8K notably shows a significant decrease in accuracy on Phi-1.5 using CoT contrast. This could be due to contamination in the Phi-1.5B dataset \citep{zhang2024careful} 

There was also contamination in the Mistral model with GMS8K, so that may have affected the results there too \citep{zhang2024careful}. 

But the results demonstrate that contrastive CoT is successful when addressing common-sense multiple choice questions. The only time that one of the amateurs does not have a greater value than the baseline on multiple choice is GPT-3.5 on AquA. Some improvements are much more modest than others, but all other model-database combinations that are muultiple choice do improve.  

The Mistral multiple-choice offers more improvement than Phi's multiple-choice which could be due to Mistral being a bigger model with more parameters. It having more parameters could allow it to use the contrasting more powerfully.

\hyperref[sec:Table 5]{Table 5} shows an interesting property of the outputs, as it appears the baseline evaluates its expressions with lower accuracy than most of the other experiments, but still results in a higher accuracy. After conducting analyses on model output for each of the experiments, it appears as though, for the experiments, the mathematical calculations may be higher in accuracy in the Amateurs, but the expressions themselves contain incorrect numerical values which may indicate an incorrect translation from the word questions to expressions.

The overall results demonstrate that contrasting with CoT prompting and amateur prompting is mostly successful when addressing common-sense multiple choice questions.

\begin{table*}[hbt!]\label{sec:Table 2}
\centering
\resizebox{0.6\textwidth}{!}{%
\begin{tabular}{l|l|l|l|l}
\toprule
\midrule
\textbf{Model} & \textbf{Dataset} & \textbf{Baseline} & \textbf{Amateur 1 ($\alpha$ = 0.5)} & \textbf{Amateur 1 ($\alpha$ = 0.8)}\\
\midrule
Phi-1.5 & GSM8K & \textbf{34.3} & 26.9 & 23.9\\
Phi-1.5 & AQuA & 19.3 & 22.1 & \textbf{25.6}\\
Phi-1.5 & CommonSenseQA & 24.5 & 24.1 & \textbf{24.9}\\
\midrule
Mistral 7B & GSM8K & 41.3 & \textbf{42.0} & 34.6\\
Mistral 7B & AQuA & 30.3 & \textbf{33.1} & 29.1\\
Mistral 7B & CommonSenseQA & 47.1 & 48.7 & \textbf{52.0}\\
\midrule
\bottomrule

\end{tabular}
}

\caption{Accuracies (in percentages) of baseline and contrastive CoT with varying $\alpha$. CCoT typically outperforms standard CoT on arithmetic and commonsense reasoning tasks.}

\label{tab:my_table}
\end{table*}

\begin{table*}[hbt!]
    \centering
    \resizebox{0.4\textwidth}{!}{%
    \begin{tabular}{|l|c|c|c|}
        \toprule
        Method & GMS8K & AQuA & CommonSense \\
        \midrule
        Baseline & \textbf{58.7} & \textbf{52.8}  & 61.4\\
        Amateur 1($\alpha$ = 0.8) & 53.9 & 28.7 & \textbf{64.4} \\
        \bottomrule
    \end{tabular}
    }
    \caption{GPT-3.5 Results on Amateur 1 on 0.8 Alpha. GPT-3.5 use was limited due to expenses so testing was only possible with one Alpha value. It does not perform as well on GPT, with only one of the CCoT having a better result.}
    \label{tab:my_table}
\end{table*}

\begin{table*}[hbt!]
\centering
\resizebox{0.8\textwidth}{!}{%
\begin{tabular}{l|l|l|l|l|l|l}
\toprule
\midrule
\textbf{Model} & \textbf{Dataset} & \textbf{Baseline} & \textbf{Amateur 2 ($\alpha$ = 0.5)} & \textbf{Amateur 2 ($\alpha$ = 0.8)} & \textbf{Amateur 3($\alpha$ = 0.5)} & \textbf{Amateur 3($\alpha$ = 0.8)} \\
\midrule
Phi-1.5 & GSM8K & \textbf{34.3} & 30.2 & 26.3 & 33.4 & 31.6\\
Phi-1.5 & AQuA & 19.3 & \textbf{23.6} & 21.3 & 22.0 & 21.7\\
Phi-1.5 & CommonSenseQA & 24.5 & 23.5 & 15.6 & 24.1 & \textbf{24.9}\\

\midrule

Mistral 7B & GSM8K & 41.3 & 38.9 & \textbf{43.2} & 42.2 & 41.6\\
Mistral 7B & AQuA & 30.3 & \textbf{35.8} & 33.9 & 31.5 & 29.9\\
Mistral 7B & CommonSenseQA & 47.1 & 26.1 & \textbf{48.2} & 26.1 & 39.2\\

\midrule
\bottomrule

\end{tabular}
}
\caption{Results on Other Amateurs. Amateur 2 seems promising because out of the six categories, it has the best result in four of them.}
\label{tab:my_table}
\end{table*}

\begin{table*}[hbt!]\label{sec:Table 5}
\centering
\resizebox{1\textwidth}{!}{%
\begin{tabular}{l|l|l|l|l|l|l|l|l|l}
\toprule
\midrule
           & \textbf{Baseline} & \textbf{Amateur 1 (0.5)} & \textbf{Amateur 1 (0.8)} & \textbf{Amateur 2 (0.5)} & \textbf{Amateur 2 (0.5)} & \textbf{Amateur 3 (0.5)} & \textbf{Amateur 3 (0.5)} & \textbf{Coherence Boosting (0.5)} & \textbf{Coherence Boosting (0.8)} \\
\midrule
Mean       & 6.110   & 6.088          & 6.039          & 6.071          & 6.037          & 6.062          & 6.089          & 4.386                   & 5.331                   \\
\midrule
Proportion & 0.589   & 0.617          & 0.626          & 0.607          & 0.615          & 0.608          & 0.606          & 0.529                   & 0.542                   \\ 

\midrule
\bottomrule

\end{tabular}
}
\caption{Comparing mean number of sentences per output and the proportion of expressions evaluated correctly for each experiment. Alpha values are in parenthesis.}
\end{table*}

\section{Related Work}

This work is inspired by three main ideas: Context-Aware Decoding, Chain of Thought, and prompting methods. Here we detail the most relevant papers in these paths.

The first direction that this paper relates to is prompting. The last few years have given way to major improvements in prompting. This is apparent with the popularity of few-shot prompting by \citep{brown2020language}, automatically learning prompts \citep{lester-etal-2021-power}, or providing model prompts by describing tasks 
\citep{wei2022finetuned,sanh2022multitask,ouyang2022training}. This idea leads to our paper's relation to Chain-of-Thought Prompting.

Chain-of-Thought prompting was a leading discovery with the ability to improve model performance significantly without scaling size. With such an improvement in performance, one of the main focuses of this paper was trying to take the improvement even further. Our main reference would be \citep{wei2023chainofthought} for its groundbreaking discovery and testing of arithmetic and common-sense benchmarks. This paper uses similar benchmarks and prompting for testing chain of thought with context-aware decoding. 

Lastly, this paper relates to others that pushed for context-aware decoding and its ability to decrease hallucination in LLMs. In particular, the paper that was referenced was \citep{shi2023trusting} for its novelty in getting LLMs to pay more attention to the context provided. The context-aware decoding was effectively used in this paper to improve the accuracy of models and decrease hallucination.

\section{Conclusion}

    Our findings highlight the impact of integrating expert and amateur prompts in Context-Aware Decoding across diverse datasets. This study highlights the need for future research prioritizing context-sensitive applications in language modeling. Further, there is a pressing need to explore the scalability of these techniques and their relevance across even broader datasets and across differing models and sizes. Future investigations could delve into testing a wider range of alpha values to unveil potentially more substantial results, aiding our understanding of the nuances within context-driven language modeling.

\section{Limitations}

Our analysis could be limited in that the difference in strength between the expert and amateur prompting is too inconsequential. That means that there is a possibility that the models themselves have generated outputs that are not dissimilar enough to impact results. 
    
Another limitation specifically on the Phi-1.5 GMS8K   results could be the fact that Phi-1.5 is possibly contaminated with GSM8K \citep{zhang2024careful}. This explains why our trials for Phi-1.5 produced many unfinished and illogical answers when responding to input prompts, yet performed much better on the baseline compared on Amateur 1 on GSM8K (Table 2). 

Phi-1.5 has also been known to work best with prompts in a QA format which inherently makes the format of the Amateur 1 and 2 prompts difficult for the model to synthesize outputs, and thus impact results \citep{textbooks2}. This could also result because of the fact that Phi-1.5 is too weak of a model with only 1.3 billion parameters to accurately display the effect of CoT contrasting.

Additionally, GPT-3.5 Turbo is a much larger model compared to Phi-1.5 and Mistral 7B. This could affect results too, because while it seems that the bigger size could make it more effective, as it's able to do contrasting more effectively, which it might, the contrasting could also get diluted in the billions of parameters.

Among the three, only GPT-3.5 Turbo is an instruction-tuned model, meaning that it's ability to reason through questions may not be a result of CoT contrasting but rather because it already contains such capabilities. There also might be factors similar to our contrasting methods already hidden in their instructions, which would make our methods less effective.

The under-performance of GPT-3.5's GSM8K and AQuA results may be attributed to GPT-3.5's difficulty in consistently answering mathematical questions accurately \citet{frieder2023mathematical}, as tested GPT versions of January 9, January 30, and GPT-4 extensively with GHOSTS, a natural language mathematics dataset, and concluded they were not competent in advanced mathematical comprehension. This could explain why the accuracy did not improve even with context-aware decoding. 

For future experiments, the first steps to improve should be to run experiments on non-instruction tuned and much larger models that can better reflect whether or not chain-of-thought vs. amateur contrasting improves the accuracy of outputs. There could also be more experiments done on different datasets, to see how the constrasting works differently for different types of problems. Additionally, selecting an optimal $\alpha$ value for each individual task, question or even token could potentially improve performance. 

\bibliography{anthology,custom}

\appendix

\end{document}